\documentclass[10pt,twocolumn,letterpaper]{article}

\usepackage[pagenumbers]{cvpr} 

\usepackage{times}
\usepackage{epsfig}
\usepackage{graphicx}
\usepackage{amsmath}
\usepackage{amssymb}
\usepackage{multirow}
\usepackage{bm}
\usepackage[bottom]{footmisc}

\usepackage{adjustbox}
\usepackage{setspace}
\usepackage[normalem]{ulem}

\usepackage{booktabs}
\usepackage{subcaption}
\usepackage{parselines}
\usepackage{lineno}
\usepackage{setspace}
\usepackage{emptypage}
\usepackage{rotating}
\usepackage{booktabs}
\usepackage{makecell}
\usepackage{algorithm,tabularx}
\usepackage{csquotes}
\usepackage{xhfill}
\usepackage{xcolor}
\usepackage{soul}
\usepackage{bbm}
\usepackage{dsfont}
\usepackage{float}
\usepackage{algpseudocode}
\usepackage{colortbl}
\usepackage[table,xcdraw]{xcolor} 

\definecolor{customcolor1}{HTML}{7DA7B4}
\definecolor{cvprblue}{rgb}{0.21,0.49,0.74}
\usepackage[pagebackref,breaklinks,colorlinks,allcolors=cvprblue]{hyperref}

\def\paperID{7611} 
\def\confName{CVPR}
\def\confYear{2025}

\title{FreqDebias: Towards Generalizable Deepfake Detection via Consistency-Driven \\Frequency Debiasing}

\author{Hossein Kashiani\qquad Niloufar Alipour Talemi\qquad Fatemeh Afghah\\
Clemson University\\
{\tt\small \{hkashia, nalipou, fafghah\}@clemson.edu
}
}

\begin{document}
\maketitle
\begin{abstract}

Deepfake detectors often struggle to generalize to novel forgery types due to biases learned from limited training data. In this paper, we identify a new type of model bias in the frequency domain, termed spectral bias, where detectors overly rely on specific frequency bands, restricting their ability to generalize across unseen forgeries. To address this, we propose FreqDebias, a frequency debiasing framework that mitigates spectral bias through two complementary strategies. First, we introduce a novel Forgery Mixup (Fo-Mixup) augmentation, which dynamically diversifies frequency characteristics of training samples. Second, we incorporate a dual consistency regularization (CR), which enforces both local consistency using class activation maps (CAMs) and global consistency through a von Mises-Fisher (vMF) distribution on a hyperspherical embedding space. This dual CR mitigates over-reliance on certain frequency components by promoting consistent representation learning under both local and global supervision. Extensive experiments show that FreqDebias significantly enhances cross-domain generalization and outperforms state-of-the-art methods in both cross-domain and in-domain settings.

\end{abstract}

\section{Introduction}

Recent advances in face forgery enable malicious users to create ultrarealistic facial images that appear genuine, which poses threats to societal trust and security \cite{jiang2020deeperforensics,li2020celeb,DFDC,DFD}. While detection methods have shown progress, they often struggle to generalize beyond specific in-domain forgery types, limiting their effectiveness in detecting out-of-distribution samples \cite{dong2022explaining,luo2021generalizing,fei2022learning,zhao2021multi,li2021frequency,huang2023implicit,chen2022self,chen2022ost,shiohara2022detecting,liang2022exploring}. This limitation arises from model bias, where detectors overly rely on spurious correlations, such as identity \cite{huang2023implicit}, background \cite{liang2022exploring}, specific-forgery \cite{LSDA_YAN_CVPR2024,Yan_2023_ICCV}, or structural information \cite{cheng2024ed}, rather than common forgery features shared across various forgeries. While recent studies \cite{Yan_2023_ICCV, Dong_2023_CVPR, huang2023implicit, LSDA_YAN_CVPR2024,cheng2024ed} have focused on addressing human-perceptible bias, to the best of our knowledge, imperceptible model bias in the frequency domain remains unexplored. An analysis in this domain can provide insights into the causes of this bias and support mitigation efforts to enhance generalization.

\begin{figure}[t] 
\centering
    \includegraphics[scale = 0.195]{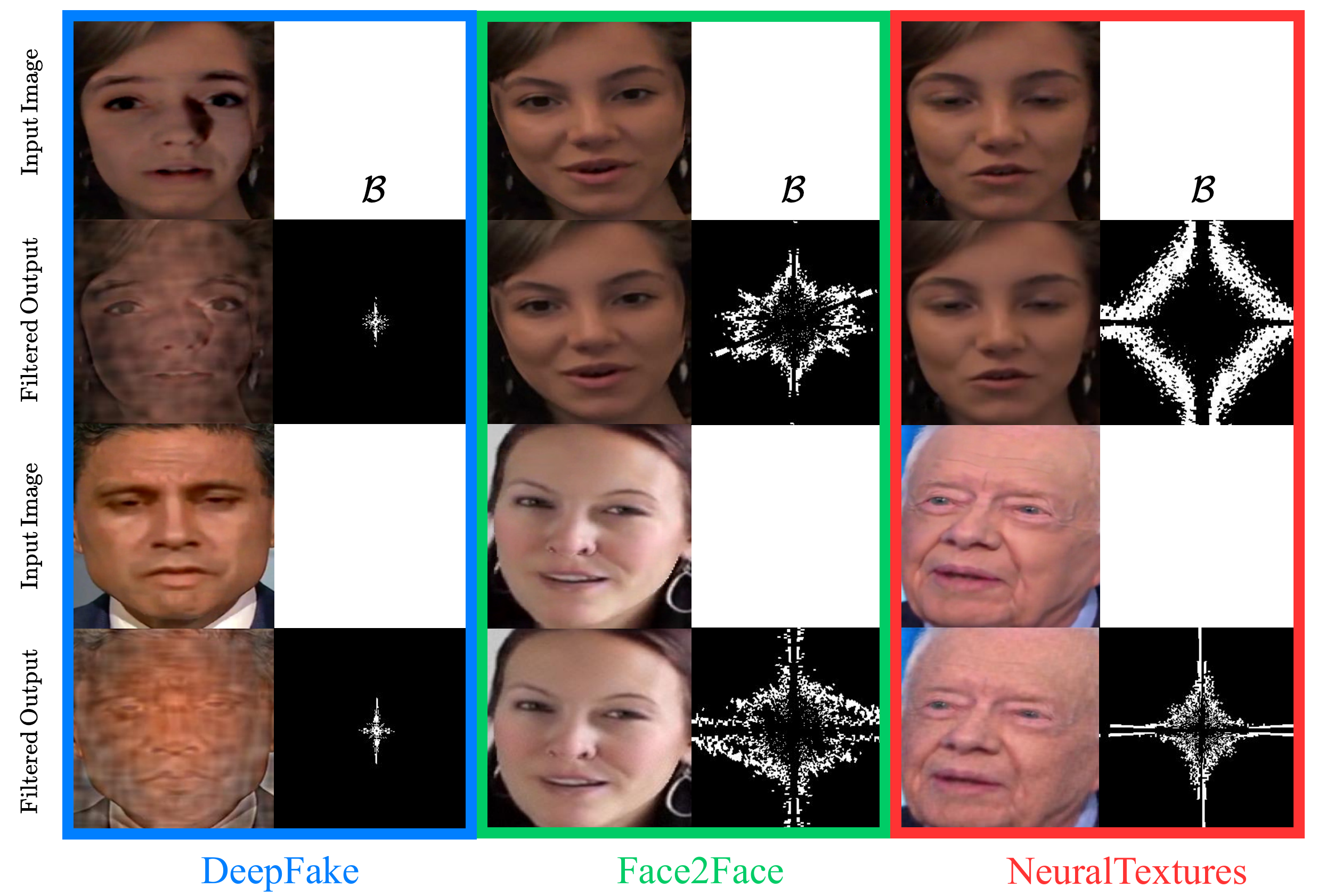}
    \caption{Visualization of certain frequency bands overly relied upon by the vanilla deepfake detector. The dominant frequency bands, highlighted in binary masks $\mathcal{B}$, are restricted to specific bands and also driven by the forgery type. Further details are provided in the supplementary material.}
    \label{mask}
\end{figure}

In this paper, we investigate model bias in the frequency domain and evaluate its impact on the generalization capabilities of deepfake detectors. To this end, we design a spectral bias experiment to identify the range of frequency bands that a standard deepfake detector pays attention to during classification. Our analysis reveals that detectors often over-rely on certain frequency bands, termed dominant frequency components, while failing to capture general forgery artifacts that may be spread across wider frequency bands. These components are defined as bands whose exclusion leads to the largest increase in classification loss, indicating the detector’s critical dependence. As shown in Fig.~\ref{mask}, these dominant frequency components are primarily driven by the forgery type, each exhibiting unique statistical patterns and typically confined to specific frequency bands. This over-reliance limits the generalization capacity of detectors across unseen forgeries. Our findings align with prior study \cite{Linear} showing that neural networks prioritize dominant frequency components in the statistical structure of training data to reduce optimization loss, at the cost of capturing semantic features necessary for generalization.

Building on these observations, we introduce a frequency debiasing framework, termed FreqDebias, to mitigate spectral bias and improve generalization of our detector beyond in-domain forgeries. FreqDebias incorporates two complementary strategies to achieve this objective. First, we introduce a novel Forgery Mixup (Fo-Mixup) which diversifies the frequency characteristics of training samples to address the over-reliance of our detector on dominant frequency components. Fo-Mixup dynamically modulates the amplitude spectra within these dominant frequency bands, creating diverse and challenging forgery samples to broaden the exposure of our detector to varied frequency bands.

Second, to explicitly regularize the learning behavior of our detector across various frequency components, we propose a dual consistency regularization (CR) that enforces our detector to learn consistent representation in both local and global supervision. For local supervision, the sensitivity of the last convolution layer w.r.t. the spatial distribution of discriminative patterns is regularized using class activation maps (CAMs) \cite{zhou2016learning}. Since Fo-Mixup preserves the spatial location of tampered regions, enforcing CAM consistency ensures that our forgery detector remains focused on the discriminative regions, making it less susceptible to domain-specific noise or other forgery-irrelevant artifacts that may arise from dominant frequency components.

Moreover, for global supervision, we model facial feature representations using the von Mises-Fisher (vMF) distribution to impose CR on a hyperspherical embedding space. Compared with the Gaussian distribution, the vMF distribution exploits the geometric properties of hyperspherical space to learn a global view of the embedding space. \cite{li2021spherical,wang2022towards,hasnat2017mises,kobayashi2021t}. Also, the vMF distribution can better capture semantic information where embedding directions specify semantic classes \cite{shaban2022few}. Leveraging this property, we impose CR on this unit hypersphere to facilitate global, geometrically consistent learning as shown in Fig.~\ref{framework}. To substantiate the generalization of our method, we conduct extensive experiments under diverse face forgery attacks. The results validate that our FreqDebias framework significantly outperforms the state-of-the-art (SOTA) studies under the cross-domain setting while maintaining its competitive performance under the in-domain setting. 

Our main contributions are summarized as follows:

\begin{enumerate}
  \item We discover spectral bias in deepfake detection and propose the Fo-Mixup augmentation to mitigate this bias.
  Fo-Mixup exposes the detector to a wider range of frequencies to reduce over-reliance on specific frequency bands, thus enhancing its generalization capabilities.

  \item We introduce a dual consistency regularization (CR) to explicitly regulate our detector by jointly enforcing local and global consistency. By leveraging CAMs for local consistency to maintain focus on discriminative regions and the vMF distribution for global consistency on a hyperspherical embedding space, our dual CR effectively mitigates the spectral bias.
  \item Extensive experiments in both in-domain and cross-domain settings demonstrate the effectiveness and generalization of our framework compared to existing studies.

\end{enumerate}

\section{Related Works}
\paragraph{Deepfake Detection.}
In response to the growing threat of face forgery attacks, numerous studies have advanced forgery detection techniques \cite{shiohara2022detecting,gu2022hierarchical,luo2021generalizing}. Some studies focus on capturing forgery artifacts in the spatial \cite{li2020face,shiohara2022detecting} or frequency \cite{luo2021generalizing,liu2021spatial} domains. For instance, Qian et al. \cite{qian2020thinking} utilize local frequency statistics via the Discrete Cosine Transform, while Liu \textit{et al.} \cite{liu2021spatial} apply the Discrete Fourier Transform to detect up-sampling artifacts. However, their performance degrades significantly on unseen forgeries. To address generalization, recent studies \cite{Yan_2023_ICCV,Guo_2023_ICCV,huang2023implicit,luo2021generalizing} propose various strategies. One method, CORE \cite{ni2022core}, addresses overfitting by enforcing representation consistency across pairs of standard augmentations to enhance generalization. However, CORE limits the potential of CR by restricting augmentation diversity. In contrast, our method synthesizes challenging forgeries to address the spectral bias in deepfake detection to learn more generalizable representations.

\vspace{1mm}

 \noindent\textbf{Debiasing and Spurious Correlations.} Recent studies have demonstrated that deep learning models are vulnerable to spurious correlations \cite{gavrikov2024can,lin2024shortcut}. These are irrelevant features linked to specific classes that harm generalization \cite{geirhos2020shortcut}. In deepfake detection, such biases emerge as dependence on unintended artifacts like identity \cite{huang2023implicit}, background \cite{liang2022exploring}, structural information \cite{cheng2024ed}, or forgery-specific patterns \cite{LSDA_YAN_CVPR2024,Yan_2023_ICCV}, degrading performance on unseen forgeries. Additionally, such correlations can emerge in the frequency domain, driving models to exploit spectral shortcuts \cite{wang2022frequency,islam2022frequency}. Data augmentation is an effective strategy to counteract these biases \cite{yao2022improving,puli2022nuisances}. While prior studies \cite{huang2021fsdr,xu2021fourier,yang2020fda, yucel2023hybridaugment,zhang2023semi} have explored Fourier-based methods, they overlook the critical role of dominant frequencies. Our method explicitly disrupts dominant frequencies to mitigate spurious shortcuts and improve generalization.

\vspace{1mm}

\noindent\textbf{Von Mises-Fisher Distribution.} The vMF distribution is a probability distribution function for directional data which can statistically model samples with unit norm. An important advantage of the vMF distribution over the Gaussian distribution is its ability to model directional data with a hyperspherical interpretation. This ability makes it more appropriate as a prior for directional data. Several recent studies in computer vision \cite{shaban2022few,hasnat2017mises,wang2022towards,li2021spherical,xu2023probabilistic} have highlighted the effectiveness of the vMF distribution in hyperspherical representation learning. In \cite{kobayashi2021t}, the vMF distribution regularizes intra-class feature distribution to improve performance on imbalanced data. Kirchhof \textit{et al.} \cite{kirchhof2022non} introduce probabilistic proxy-based deep metric learning using directional vMF distributions to handle image-intrinsic uncertainties. Wang \textit{et al.} \cite{wang2022towards} extend cosine-based classifiers with a vMF mixture model, enabling a quantitative quality measurement in hypersphere space for long-tailed learning. In this study, we utilize the vMF distribution to globally enhance CR across diverse face forgery variations, ensuring robust generalization.

\section{Methodology}

Given an original training dataset \( T = \{(x_i^{t}, y_i^{t})\}_{i=1}^{N_t} \) with \( N_t \) samples, where each \( x_i^{t} \) and \( y_i^{t} \) represent the \( i \)-th image and its corresponding class label, we employ Fo-Mixup to generate a diverse set of synthetic forgery images. Additionally, we use standard data augmentation to produce augmented versions of the real images. This process yields a synthesized dataset \( S = \{(x_i^{s}, y_i^{s})\}_{i=1}^{N_s} \) comprising \( N_s \) samples, where \( x_i^{s} \) and \( y_i^{s} \) denote the \( i \)-th synthetic image and its class label, respectively. Using both the synthesized forgery images and the augmented real images, we then regularize our model to maintain consistency across the original \( T \) and synthesized \( S \) domains for both real and forgery classes.

\subsection{Forgery Mixup Augmentation}


The amplitude spectrum of an image retains low-level statistics, such as forgery artifacts, color, style information, and sensor artifacts, while the phase spectrum includes high-level semantic content \cite{xu2021fourier,piotrowski1982demonstration,oppenheim1981importance}. In deepfake detection, training the detector without specific considerations can cause the model to over-rely on domain-specific details (such as hair color, background, texture, and style) encoded in the amplitude spectrum \cite{luo2021generalizing}. Our experiments show that this overemphasis on dominant frequency components can lead to overfitting, where the model performs well on training data but poorly on out-of-distribution forgeries.

 To address these challenges, we introduce the Fo-Mixup augmentation to modulate the dominant frequency components within the amplitude spectrum of different forgery images. Fo-Mixup exposes our forgery detector to more variations in the dominant frequency components of the amplitude spectrum during training. This exposure encourages the detector to attend to a broader range of frequency bands, thereby promoting the learning of more generalizable features and reducing overfitting.

\begin{algorithm}[t]
\caption{Fo-Mixup Augmentation}
\label{alg:fomixup}
\begin{algorithmic}[1]
\footnotesize
\setstretch{1} 

\Statex \textbf{Inputs:} Input forgery images $x_i$ and $x_j$, interpolation hyperparameter $\xi$, total number of angular segments $T$, total number of clusters $k$, top number of clusters $t$, constant distance $r$, constant angle $\theta$ from the horizontal axis, a normal distribution $p_{\mathcal{A}}$, pre-trained forgery \mbox{detector}.
\Statex \textbf{Outputs:} Augmented forgery image $x_{ij}$\\

$[\mathcal{A}(x_i), \mathcal{P}(x_i)] \leftarrow \mathcal{F}(x_i)$ \Comment{ FFT for $x_i$}\\

$[\mathcal{A}(x_j), \mathcal{P}(x_j)] \leftarrow \mathcal{F}(x_j)$ \Comment{ FFT for $x_j$}\\

$\{\mathcal{A}_{\text{seg}_1},\ldots,\mathcal{A}_{\text{seg}_T} \} \leftarrow \operatorname{Partition}(\mathcal{A}(x_i,r,\theta))$  \Comment{ Partition $\mathcal{A}(x_i)$ into $T$ angular segments with constant distance $r$ and angle $\theta$}\\

$ \{\mu_{\text{seg}_1}, \ldots, \mu_{\text{seg}_T}\} \leftarrow \{\frac{1}{{N}_{\text{seg}_z}} \sum_{\mathcal{A}_{\text{seg}_z}} \log(1+|\mathcal{A}_{\text{seg}_z}|)\}^{T}_{z=1}$\newline\Comment{Mean log-scaled spectrum}\\%

 $\{\mu_1, \ldots, \mu_k\} \leftarrow \operatorname{K-means}(\{\mu_{\text{seg}_1}, \ldots, \mu_{\text{seg}_T}\})$\Comment{Clustering}\\


$\{\mathcal{B}_{1},\ldots,\mathcal{B}_k\} \leftarrow \{ \text{indices}_{\mu_z}\}^{k}_{z=1}$ \Comment{Binary masks}\\

 $\{\mathcal{A}_{\mu_1}, \ldots, \mathcal{A}_{\mu_k}\}\leftarrow \{\mathcal{A}(x_i)\otimes\mathcal{B}_{z}\}^{k}_{z=1}$ \Comment{Filtering}\\

 $\{x_{{\mu_1}}, \ldots, x_{{\mu_k}}\}\leftarrow \{\mathcal{F}^{-1}[\mathcal{\mathcal{A}}_{\mu_z},\mathcal{P}(x_i)]\}^{k}_{z=1}$ \Comment{Filtered images}\\

$\{x_{{\mu_{s1}}}, \ldots, x_{{\mu_{st}}}\} \leftarrow \operatorname{OHEM}(\{x_{{\mu_1}}, \ldots, x_{{\mu_k}}\})$ \Comment{Apply OHEM \cite{shrivastava2016training} to the filtered images using our pre-trained forgery detector}\\
 
$\{\mathcal{B}_{1},\ldots,\mathcal{B}_t\} \leftarrow \{ \text{indices}_{\mu_{sz}}\}^{t}_{z=1}$ \Comment{Keep top $t$ clusters, selected by OHEM in step 9. These clusters correspond to the highest loss values.}\\

$\mathcal{B} \leftarrow \operatorname{RandomSelect}(\mathcal{B}_{1},\ldots,\mathcal{B}_t)$ \Comment{Random selection}\\

$\mathcal{\hat{A}}(x_{ij})\leftarrow\mathcal{A}(x_i)\otimes\mathcal{B} +\left[(1-\xi) \mathcal{A}(x_i) +\xi \mathcal{A}(x_j)\right]\otimes(1-\mathcal{B})$ \\

$x_{ij}\leftarrow \mathcal{F}^{-1}\left[p_{\mathcal{A}}\otimes\mathcal{\hat{A}}(x_{ij}),\mathcal{P}(x_i)\right]$\Comment{Reconstruct image}
\State \textbf{Return  $x_{ij}$} \Comment{Augmented forgery image}
\end{algorithmic}\label{alg:fomixup}
\end{algorithm}

\begin{figure*}[t] 
\centering
\hspace{2cm} 
\includegraphics[width=0.99\textwidth]{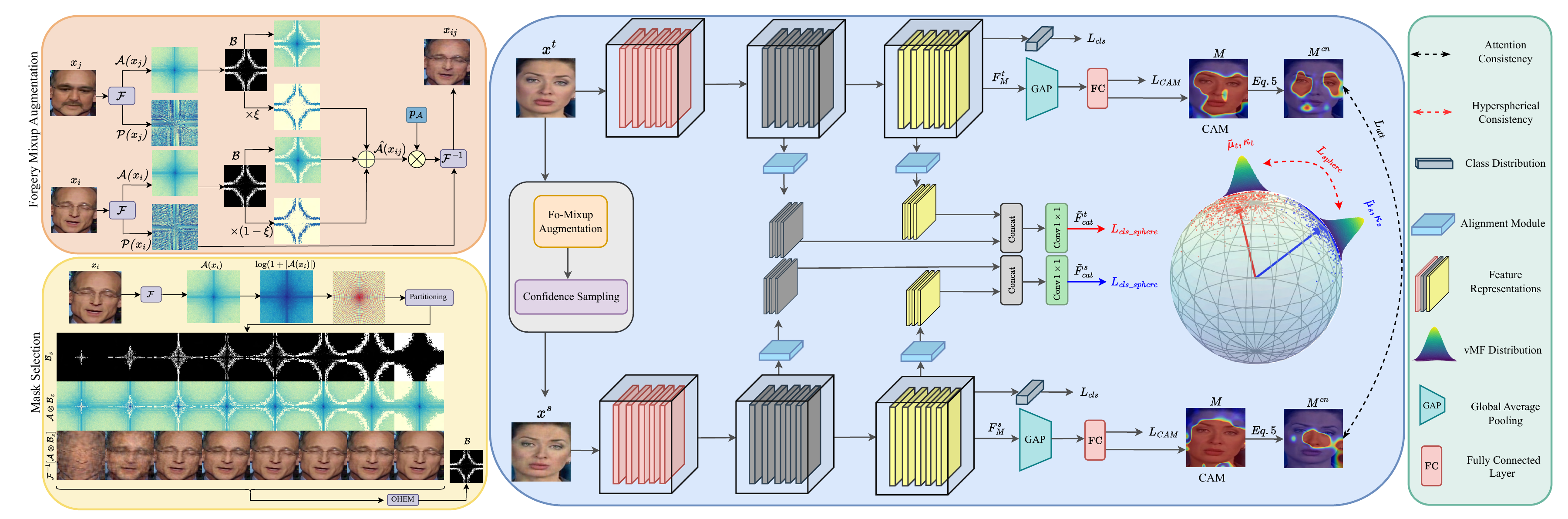}
\caption{Overview of the proposed FreqDebias framework. First, Fo-Mixup diversifies the forgery class across frequency spectra by transforming $x^t$ into $x^s$, while preserving the spatial structure of tampered regions. Then, the network is jointly regularized at local and global levels to enforce consistent representation learning. For the real class, the same process is applied using standard augmentation.}
\label{framework}
\end{figure*}

Given an input forgery image $x_i$, we apply the fast Fourier transformation $\mathcal{F}(x_i)$ on each RGB channel to compute its amplitude $\mathcal{A}(x_i)$ and phase $\mathcal{P}(x_i)$ components. Fo-Mixup initiates with the identification of the key frequency components within the input forgery image.

To achieve this, we partition the amplitude spectrum into angular segments, with radial and angular components divided into intervals \( r \in [i_r \Delta r, (i_r+1) \Delta r] \) and \( \theta \in [i_\theta \Delta\theta, (i_\theta+1) \Delta\theta] \). The radial component measures distance from the origin, while the angular component defines orientation in the frequency domain. The indices \( i_r \) and \( i_\theta \) denote discrete steps, and the step sizes \( \Delta r \) and \( \Delta \theta \) determine the segmentation resolution. For each segment, we compute the mean logarithmic-scaled spectrum denoted as $\mu_{\text{seg}}$, as the average of $\log(1+|\mathcal{A}_{seg}|)$ values within the segment, where $\mathcal{A}_{seg}$ is the segment’s amplitude spectrum.

Next, we apply the K-means clustering algorithm \cite{macqueen1967some} to all angular segments using the mean spectrum $\mu_{\text{seg}}$ of all angular segments. The $k$ cluster indices are then employed to construct $k$ binary masks, which are subsequently used to filter the input image into $k$ distinct filtered images. We feed the filtered images into our detector and arrange the classification losses in descending order, following the Online Hard Example Mining (OHEM) \cite{shrivastava2016training}. Finally, we randomly select one of the top $t$ binary masks (denoted as $\mathcal{B}$) as the ones associated with the highest loss values. To integrate the key frequency components of different forgery images, the selected binary mask $\mathcal{B}$ is applied to the amplitude component (shown in Fig. \ref{framework}). These operations are formulated as follows:
\begin{multline}\label{eq.4}
\mathcal{\hat{A}}(x_{ij})=\mathcal{A}(x_i)\otimes\mathcal{B}  
\\ +\left[(1-\xi) \mathcal{A}(x_i) +\xi \mathcal{A}(x_j)\right]\otimes(1-\mathcal{B}),
\end{multline}
\noindent where $x_i$ and $x_j$ are the randomly selected forgery inputs, and $\xi$ denotes the hyperparameter that adjusts the proportion of inputs. Also, $\otimes$ denotes element-wise multiplication. To expose our model in CR to more forgery artifacts variations in the dominant frequency component, we also perturb the amplitude spectrum as $p_{\mathcal{A}}\otimes\mathcal{\hat{A}}(x_{ij})$, where $p_{\mathcal{A}}$ is a normal distribution $ \mathcal{N}(1,0)$. Finally, the inverse Fourier transform reconstructs the newly stylized image using the phase component of the original input $\mathcal{P}(x_i)$ and the integrated amplitude spectrum $\mathcal{\hat{A}}(x_{ij})$ as follows:
\begin{equation}\label{eq.5}
\begin{aligned}
x_{ij}= \mathcal{F}^{-1}\left[\left(p_{\mathcal{A}}\otimes\mathcal{\hat{A}}(x_{ij})\right)* e^{-i*\mathcal{P}(x_i)(u,v)}\right],
\end{aligned}
\end{equation}
\noindent where $u$ and $v$ represent coordinates in frequency domain. The generated image $x_{ij}$ retains the semantic content of $x_i$ with diversified dominant frequency components. This expands the detector focus across frequency bands. The Fo-Mixup augmentation is outlined in Algorithm \ref{alg:fomixup}.

\vspace{1mm}
\noindent\textbf{Confidence Sampling.}
Augmented samples with high prediction entropy (i.e., low confidence) hinder generalization performance. To address this, we filter out unreliable samples from the pool of augmented data using Shannon entropy, defined as \( E_{x_{ij}} = -\mathbf{p}(x_{ij}) \log \left(\mathbf{p}(x_{ij})\right) \), where \( x_{ij} \) is the augmented sample and \( \mathbf{p}(x_{ij}) \) denotes the model’s output probability for that sample. After calculating the entropy for each prediction, we rank the samples in ascending order and select the top \( \lambda \)  percent of high-confidence samples.

\subsection{Consistency Regularization}

Our model consists of a feature extraction module with multiple stages, represented as $\Psi = \psi_M \circ \dots \circ \psi_2 \circ \psi_1$, where $M$ denotes the number of stages in the network. Since each layer captures concepts at different abstraction levels \cite{zhang2022attributable}, from low-level textures to high-level semantics, we introduce a feature alignment module $\omega_i$ after each stage to capture regularized semantic abstractions progressively across the network. We express these as $F_i = \omega_i \circ \psi_i(x) \circ \dots \circ \psi_1(x)$, where $i \in {1, \dots, M}$. The feature alignment module, which includes convolutional and residual blocks similar to those in ResNet \cite{he2016deep}, ensures that each auxiliary branch undergoes the same number of down-sampling operations as the backbone, supporting a fine-to-coarse feature transformation. With the feature alignment modules in place, the backbone is trained and its class posterior distributions for the predictions of the synthesized $(x^{s}, y) \in {S} $ and training images $(x^{t}, y) \in {T} $ are regularized as follows:
\begin{multline}\label{eq.10}
    L_{{cls}} = \mathcal{L}_{{CE}} \Bigl( \sigma\bigl(\alpha_{f} (x^s)\bigr), y \Bigr) + \mathcal{L}_{{CE}} \Bigl( \sigma\bigl(\alpha_{f} (x^t)\bigr), y \Bigr) \\
    + {D_{KL}} \Bigl( \sigma\bigl(\alpha_{f}(x^s); \tau\bigr), \sigma\bigl(\alpha_{f}(x^t); \tau\bigr) \Bigr),
\end{multline}
\noindent where $\mathcal{L}_{{CE}}$ indicates the standard cross-entropy (CE) loss function, $D_{{KL}}$ denotes the Kullback-Leibler (KL) divergence, and $\alpha_{f} (x)$ is the final prediction of our model. Also, ${\sigma}(\alpha(x) ; \tau)$ denotes the Softmax operation with temperature $\tau$. When $\tau=1$, it would be a normal Softmax operation, expressed as ${\sigma}(\alpha_i (x))$. 

\vspace{1mm}
\noindent\textbf{Attention Consistency Regularization.} With the synthesized forgery samples covering a wider range of dominant frequency components, we first regularize our detector using local consistency to identify consistent visual cues, regardless of dominant frequency variations. To achieve this, we compute CAMs \cite{sun2020fixing,zhou2016learning} on the feature map \( F_M \in \mathbb{R}^{c \times h \times w} \) at the final stage of our backbone, where \( c \), \( h \), and \( w \) represent the number of channels, height, and width, respectively. CAMs offer valuable interpretability by highlighting the critical regions in an image that are relevant for detecting forgeries. The proposed Fo-Mixup augmentation is designed to maintain the spatial location of tampered regions consistently between the input image \( x^t \) and the synthesized forgery \( x^s \), preserving critical information for detection. By enforcing CAM consistency, we ensure that our forgery detector remains focused on the discriminative regions, making it less susceptible to domain-specific noise or artifacts that could otherwise mislead its predictions. For consistency supervision at the attention level, we calculate CAMs using a global average pooling (GAP) layer followed by a fully connected (FC) layer with weights \( W \in \mathbb{R}^{c \times n} \), where \( n \), the number of classes, is two. The FC layer is trained using the CE loss function, \( L_{CAM} \), to distinguish between forgery and real samples. The CE loss is defined as \( L_{CAM} = -\sum_{i=1}^{n} y_i \log(\hat{y}_i) \), where \( y_i \) is the true label and \( \hat{y}_i \) is the predicted probability for class \( i \). The classifier weights \( W \) are convolved with the feature map \( F_M \) to compute the CAMs as follows:
\begin{equation}\label{eq.12a}
M = W^\top F_M,
\end{equation}
\noindent where \( M \in \mathbb{R}^{n \times h \times w} \). This process ensures that the CAMs can effectively highlight the most discriminative regions for the forgery classification. The CAMs are then normalized within the class region so that the highly valued features are solely assigned to the class region. For class-wise normalization, the CAMs are average pooled and clustered into different clusters using the K-means algorithm \cite{macqueen1967some}. The first cluster (formulated as ${Mask^{high}}$) highlights the class region and the remaining ones would be discarded. Then, the obtained category region with $Mask^{high}$ is further normalized by instance normalization \cite{ulyanov2017improved} to have a standard distribution. These operations are summarized as follows:
\begin{equation}\label{eq.13}
	{M^{cn}}= (M \otimes {Mask^{high}})\gamma + \beta,
\end{equation}
\noindent where ${M^{cn}}$ and ${Mask^{high}}$ denote the class-wise normalized CAMs and the highlighted class region at the last layer of our backbone. $\gamma$ and $\beta$ are learnable affine parameters in the instance normalization. After all, we match the refined CAMs (i.e., ${M^{cn}}$) between the synthesized and original training domains by the Jensen-Shannon Divergence (JSD) as follows:
\begin{equation}\label{eq.14}
	{L_{att}}= D_{JS}\Bigl({\sigma}\bigl({M^{cn}}(x^{s}); \tau\bigr) ,{\sigma}\bigl({M^{cn}}(x^{t}); \tau\bigr)\Bigr),
\end{equation}
\noindent where ${M^{cn}}(x^{s})$ and ${M^{cn}}(x^{t})$ are the class-wise normalized CAMs for the synthesized and training images.

\noindent\textbf{Hyperspherical Consistency Regularization.} 
Recent research demonstrates the effectiveness of hyperspherical latent spaces over Euclidean spaces in modeling the complex geometric structure of facial feature representations \cite{wang2022towards,kirchhof2022non,hasnat2017mises,li2021spherical}. To leverage this, we introduce a vMF classifier, implemented as a mixture model with two vMF distributions, to statistically model facial feature representations on the hyperspherical space for CR. To capture a global view of the embedding space for consistent geometric learning, we leverage hierarchical feature maps from different stages of the backbone network to construct our vMF classifier. To achieve this, feature maps from these stages are concatenated and resized as $\boldsymbol{F}_{cat} = \operatorname{Conv}\left(\operatorname{Concat}([F_1,....,F_M])\right)$, where $\operatorname{Conv}$ denotes the $1\times1$ convolutional layer that reduces the channel dimension of the concatenated feature maps. Given a normalized facial feature representation $\tilde{ \boldsymbol{F}}_{cat}=\boldsymbol{F}_{cat} /\left\|\boldsymbol{F}_{cat}\right\|_2$, the probability density function (PDF) for $i$-th class is formulated as follows:
\begin{gather}
p\left(\tilde{ \boldsymbol{F}}_{cat} \mid {\kappa_i}, \tilde{ \boldsymbol{\mu}}_i\right)=\mathcal{C}_d\left(\kappa_{i}\right) \exp \left({{\kappa_i} \cdot \tilde{ \boldsymbol{F}}_{cat} \tilde{ \boldsymbol{\mu}}_i^{\top}}\right),  \label{eq.15} \\
\mathcal{C}_d\left({\kappa_{i}}\right)=\frac{\kappa_i^{\frac{d}{2}-1}}{(2 \pi)^{\frac{d}{2}} \cdot I_{\frac{d}{2}-1}\left({\kappa_i}\right)}, \label{eq.16}
\end{gather}
\noindent where $\tilde{ \boldsymbol{F}}_{cat} \in \mathbb{S}^{d-1}$ is the d-dimensional unit vector, and ${\kappa_i} \in \mathbb{R}_{\geq0}$ and $\tilde{ \boldsymbol{\mu}}_i \in \mathbb{S}^{d-1}$ are trainable compactness parameter and orientation vector of the vMF classifier for $i$-th class. Also, $I_\beta (\kappa)$ is the modified Bessel function of the first kind at order $\beta$ and $C_d({\kappa})$ is a normalization constant. Given the PDF in Eq. \ref{eq.15}, and the posterior probability, the vMF classifier is trained using CE loss as below:
\begin{equation}
    	{{L}_{{cls\_sphere}}} = \mathcal{L}_{CE}\left(p(y_i \mid \tilde{\boldsymbol{F}}_{{cat}})\right), \label{eq.17}
\end{equation}
\begin{equation}
	\hfill p\left(y_i \mid \tilde{\boldsymbol{F}}_{{cat}}\right) = \frac{n_i \cdot p\left(\tilde{\boldsymbol{F}}_{{cat}} \mid {\kappa_i}, \tilde{\boldsymbol{\mu}}_i\right)}{\sum_{j=1}^2 n_j \cdot p\left(\tilde{\boldsymbol{F}}_{{cat}} \mid {\kappa_j}, \tilde{\boldsymbol{\mu}}_j\right)}, \label{eq.18}
\end{equation}
\noindent where $n_i$ indicates the cardinality of class $i$, and $p\left(y_i \mid \tilde{ \boldsymbol{F}}_{cat}\right)$ denotes the probability of $\boldsymbol{F}_{cat}$ being associated with class $i$. To enforce the geometric CR, the Distribution Matching Score ($\mathrm{DMS}$) is formulated  as follows:

\begin{equation}
\mathrm{DMS} = \left. 1 \middle/ \left( 1 + D_{\mathrm{KL}}\big( p(\tilde{ \boldsymbol{F}}_{cat}^{s} \mid \kappa_s, \tilde{ \boldsymbol{\mu}}_s),\; p(\tilde{ \boldsymbol{F}}_{cat}^{t} \mid \kappa_t, \tilde{ \boldsymbol{\mu}}_t) \big) \right) \right.
\end{equation}

\noindent where $0<\mathrm{DMS}<1$ represents the overlap between the synthesized $S$ and training domains $T$. When two domains are highly overlapped, $\mathrm{DMS} \to 1$. Finally, to enhance CR, we maximize the $\mathrm{DMS}$ between the synthesized $S$ and training domains $T$ in vMF optimization as below:
\begin{equation} \label{eq.20}
{L}_{sphere}=\mathbb{E}\left[1-\mathrm{DMS}(\boldsymbol{F}_{cat}^{s},\boldsymbol{F}_{cat}^{t})\right],
\end{equation} 

\noindent where $\mathbb{E}$ is the average function.

\vspace{1mm}
\noindent\textbf{Overall Loss.}
The overall objective function in our training optimization boils down to five loss functions as follows:
\begin{multline}\label{eq.21}
L_{total}= {L_{cls}} + \eta  {L_{CAM}}   \\ +\delta{L_{att}} +\mu{{L}_{cls\_sphere}} +\rho{L}_{sphere},
\end{multline} 
\noindent where $\eta$, $\delta$, $\mu$, and $\rho$ are hyperparameters that balance the impact of different loss functions. Having trained our proposed model end-to-end, we remove all introduced auxiliary classifiers during testing, incurring no additional computational cost compared to the baseline backbone.
 
\section{Experiments}
\subsection{Experimental settings}

\noindent\textbf{Forgery Detection Datasets.} For simplicity and fair comparison, we pursue recent deepfake detection studies \cite{DeepfakeBench_YAN_NEURIPS2023,UCF_YAN_ICCV2023,LSDA_YAN_CVPR2024,cheng2024can} and train our model on the FaceForencis++ (FF++) dataset \cite{rossler2019faceforensics}. FF++ is the widely-used video dataset with 1,000 original videos and 4,000 fake videos which are generated by four face manipulation methods, including DeepFake (DF) \cite{Deepfakes_2019}, Face2Face (FF) \cite{thies2016face2face}, FaceSwap (FS) \cite{FaceSwap}, and NeuralTextures (NT) \cite{thies2019deferred}. In terms of quality, three compression levels are utilized in FF++ generation, which are raw quality, high quality (HQ), and low quality (LQ). The HQ level is adopted to sample 32 frames in each video. To gauge the generalization and robustness ability of our model, we conduct extensive experiments on FF++, Deepfake Detection Challenge (DFDC) \cite{DFDC}, DeepFake Detection Challenge Preview (DFDCP) \cite{DFDC}, Deepfake Detection (DFD) \cite{DFD}, Celeb-DF-v1 (CDFv1) \cite{li2020celeb}, and Celeb-DF-v2 (CDFv2) \cite{li2020celeb} datasets.\vspace{1mm}\\ 
\noindent\textbf{Implementation detail.} For preprocessing and training, we adhere to the configurations outlined in DeepFakeBench \cite{DeepfakeBench_YAN_NEURIPS2023} to maintain a fair comparison, with an input size of $256 \times 256$ pixels. ResNet-34 \cite{he2016deep} is chosen as our backbone, which is initialized with the pre-trained ImageNet weights. To apply OHEM to the filtered images in Algorithm \ref{alg:fomixup}, our forgery detector is pre-trained on the FF++ dataset using the CE loss function for 30 epochs with a batch size of 32. In the CR for the forgery class, we incorporate the Fo-Mixup augmentation along with standard augmentations like contrast, pixelation, and the real class CR undergoes only standard augmentations. Our framework is trained for 50 epochs, with a batch size of 32, and an initial learning rate of $1e-4$. The hyperparameters used in Eq. \ref{eq.10} and \ref{eq.21}, Fo-Mixup augmentation, and confidence sampling are set to $\tau =4 $, $\eta =0.5$, $\delta=0.1$, $\mu=1$, $\rho=0.1$,  $k=8$,  $t=3$ , and $\lambda=0.5$. Also, $\xi$ in Eq. \ref{eq.4} is randomly sampled within $[0.0, 1.0]$. To benchmark our method, we follow the deepfake detection studies \cite{DeepfakeBench_YAN_NEURIPS2023,UCF_YAN_ICCV2023,LSDA_YAN_CVPR2024,cheng2024can} and adopt the frame-level area-under-the-curve (AUC), and Equal Error Rate (EER) metrics.

\begin{table}[t!]
    \centering
    \setlength{\tabcolsep}{2pt} 
    \resizebox{.48\textwidth}{!}{%
    \begin{tabular}{l|c@{}|ccccc|c}
        \hline
        \hline
        \multirow{2}{*}{Method} & \multicolumn{1}{c|}{In-domain} & \multicolumn{6}{c}{Cross-domain}  \\
        \cline{2-8}
        & FF++ & CDFv1 & CDFv2 &DFD& DFDCP & DFDC & C-Avg. \\
        \hline
        Xception \cite{chollet2017xception} & 96.4 & 77.9 & 73.7 & 81.6&73.7 & 70.8 & 75.54 \\
        Meso4 \cite{afchar2018mesonet} & 60.8 & 73.6 & 60.9 &54.8 &59.9 & 55.6 & 60.96 \\
        Capsule \cite{nguyen2019capsule} & 84.2 & 79.1 & 74.7 & 68.4&65.7 & 64.7 & 70.52\\
        X-ray \cite{li2020face} & 95.9 & 70.9 & 67.9 & 76.6&69.4 & 63.3 & 69.62 \\
        FFD \cite{dang2020detection} & 96.2 & 78.4 & 74.4 &80.2& 74.3 & 70.3 &75.52  \\
        F3Net \cite{qian2020thinking} & 96.4 & 77.7 & 73.5 &79.8& 73.5 & 70.2 &74.94 \\
        SPSL \cite{liu2021spatial} & 96.1 & 81.5 & 76.5 &81.2& 74.1 & 70.4 & 76.74\\

        SRM \cite{luo2021generalizing} & 95.8 & 79.3 & 75.5 &81.2& 74.1 & 70.0 & 76.02 \\
        CORE \cite{ni2022core}  & 96.4 & 78.0 & 74.3 & 80.2&73.4 & 70.5 & 75.28 \\
        RECCE \cite {cao2022end}  & 96.2 & 76.8 & 73.2 &81.2& 74.2 & 71.3 &75.34  \\
        SLADD \cite{chen2022self} & 96.9 & 80.2 & 74.0 &80.9& 75.3 & 71.7 &76.42  \\
        IID \cite{huang2023implicit} & 97.4 & 75.8 & 76.9 &79.3& 76.2 & 69.5 &75.54 \\
        UCF \cite{Yan_2023_ICCV} & 97.1 & 77.9 & 75.3 &80.7 &75.9 & 71.9 & 76.34 \\

                LSDA \cite{LSDA_YAN_CVPR2024}& -&86.7& {83.0} & \textbf{88.0}&81.5 &73.6 &82.56\\

        \hline
      FreqDebias (Ours) & \textbf{97.5} &\textbf{87.5}  &\textbf{83.6 }& {86.8 } &\textbf{82.4 } &\textbf{74.1 } &\textbf{82.88} \\
        \hline
        \hline
    \end{tabular}%
    }
    \caption{Performance comparison in both in-domain and cross-domain settings. All detectors are trained on the FF++ (HQ) dataset \cite{rossler2019faceforensics}, and evaluated on other datasets using the \textbf{frame-level} AUC metric. ``C-Avg." represents the average of all cross-domain results, with the best-performing values highlighted in bold.}
    \label{both_compariosn}
\end{table}

\vspace{1mm}
\subsection{Comparison with previous forgery methods}
\noindent\textbf{Overall Performance Across Diverse Datasets.} To explore the generalization capability of the proposed method, we train our model on the FF++ (HQ) dataset and gauge its generalization performance over other forgery datasets using the {frame-level} AUC metric. As reported in Table \ref{both_compariosn}, it is observed that the SOTA approaches experience pronounced performance drops on the cross-domain evaluations. However, the FreqDebias framework mainly generalizes better to novel forgeries and achieves SOTA performance. More precisely, the FreqDebias framework yields gains of $0.8\%$, $0.6\%$, $0.9\%$, and $0.5\%$ in terms of AUC metric compared to the runner-up methods in the CDFv1, CDFv2, DFDC, and DFDCP test sets, respectively. From this experiment, we can deduce that the proposed FreqDebias framework substantially improves the generalization of the forgery detector to unseen forgery attacks. Also, as reported in column FF++ in Table \ref{both_compariosn}, we yield great performance in the in-domain setting, compared to SOTA studies.

\vspace{1mm}
\noindent\textbf{Cross-Manipulation Evaluations.} As new manipulation techniques continuously emerge, our method should generalize well to unseen forgeries. To assess this, we follow the training scheme from DCL \cite{sun2022dual} and perform a fine-grained cross-manipulation evaluation on the FF++ dataset. Specifically, our forgery detector is tested on individual manipulation techniques while being trained on data from other techniques within the same dataset. As represented in Table \ref{Table3}, our proposed method achieves generalization improvements across four unseen forgery types, with improvements ranging from $0.08\%$ to $3.85\%$. This experiment demonstrates that our learned representation can effectively distinguish forgery artifacts from real ones in different forgeries. More experimental comparisons using the video-level AUC metric for both cross-domain and in-domain settings are provided in the \textbf{supplementary material.}

\begin{table}
\center
    \setlength{\tabcolsep}{10pt} 

\resizebox{.46\textwidth}{!}{%
\begin{tabular}{l|c|c|c|c|c}
\hline \hline 
Methods  & Train  & DF&  F2F & FS & NT\\
\hline

GFF \cite{luo2021generalizing} &  \multirow{5}{*}{ DF }   
 &99.87&  76.89 &47.21 & 72.88  \\

DCL \cite{sun2022dual}& &\textbf{99.98}& 77.13&61.01& 75.01\\ 
IID  \cite{huang2023implicit} &&99.51&-&63.83&-\\ 

SFDG \cite{Wang_2023_CVPR}&  & 99.73 &86.45 &75.34 &86.13\\ 

  FreqDebias (Ours)& 
 &  99.82  &\textbf{88.10}& \textbf{75.92} &\textbf{88.45}  \\
 \hline
  GFF \cite{luo2021generalizing} &  \multirow{4}{*}{ F2F }   
 & 89.23 &  99.10 &61.30&  64.77 \\

 DCL \cite{sun2022dual}&&91.91 &99.21& 59.58&66.67 \\ 
SFDG \cite{Wang_2023_CVPR}&  & 97.38 &99.36 &73.54& 72.61\\ 

  FreqDebias (Ours)&  
 &\textbf{98.41}&\textbf{99.44 }&\textbf{74.37} &\textbf{76.46}  \\
 \hline
  GFF \cite{luo2021generalizing} &    \multirow{5}{*}{ FS }  
 & 70.21&   68.72 &99.85&   49.91 \\

 DCL \cite{sun2022dual}&&74.80& 69.75 &\textbf{99.90}& 52.60 \\ 

 IID \cite{huang2023implicit} &&75.39 &-&99.73 &-\\ 

 SFDG \cite{Wang_2023_CVPR}&  & 81.71 &77.30& 99.53 &60.89\\ 

  FreqDebias (Ours)&  
 &\textbf{83.76} &\textbf{78.93}&99.78&\textbf{63.48}   \\
 \hline
 GFF \cite{luo2021generalizing} &      \multirow{4}{*}{ NT }  
 &88.49& 49.81 &74.31&  98.77  \\
 DCL \cite{sun2022dual}&&91.23& 52.13 &79.31& 98.97 \\ 
 SFDG \cite{Wang_2023_CVPR}&  &  91.73 &70.85 &\textbf{83.58}& 99.74\\ 
FreqDebias (Ours)& &\textbf{92.35} &\textbf{ 74.61}& {83.24}  & \textbf{99.83} \\
\hline
\hline
\end{tabular}}
\caption{Cross-manipulation performance comparison of the proposed FreqDebias framework with SOTA studies on the FF++ test set. The results are based on the \textbf{video-level} AUC metric.}
\label{Table3}%
\end{table}%

\vspace{1mm}
\noindent\textbf{Robustness Evaluations.} To evaluate the robustness of our method against diverse unseen corruptions, we follow the LipForensics \cite{haliassos2021lips} setup with uncompressed raw FF++ data. We exclude shared augmentations in our CR  to ensure a fair comparison. As shown in Table \ref{TableR}, our method demonstrates outstanding robustness, outperforming RealForensics \cite{haliassos2022leveraging} and CADDM \cite{Dong_2023_CVPR} with an impressive $97.6\%$ average AUC.

\begin{table}
\center

\resizebox{0.48\textwidth}{!}{%
\begin{tabular}{l|cccccc|cc}
\hline \hline 
Model & Saturation & Contrast& Block & Noise & Blur & Pixel&Avg \\
\hline 
Face X-ray \cite{li2020face} & 97.6 & 88.5 & 99.1 & 49.8 & 63.8 & 88.6 &81.2
 \\
     
\hline
LipForensices \cite{haliassos2021lips}& \textbf{99.9} &\uline{99.6} & 87.4 & 73.8 & 96.1 &95.6& 92.1\\

\hline

RealForensics \cite{haliassos2022leveraging} &\uline{99.8} &\uline{99.6}&98.9&79.7&95.3&98.4&95.2\\
\hline

CADDM \cite{Dong_2023_CVPR}&99.6 &\textbf{99.8} &\textbf{99.8 }&\uline{87.4} &\textbf{99.0} &\uline{98.8}& \uline{97.4} \\
\hline

FreqDebias (Ours)&99.6&\textbf{99.8}&\uline{99.7}&\textbf{89.2}&\uline{98.2}&\textbf{99.1}&\textbf{97.6}\\
\hline \hline

\end{tabular}}
\caption{Robustness evaluation in terms of \textbf{video-level} AUC on FF++ dataset. \textbf{Bold} and \uline{underline} indicate the top and second-best robustness, respectively. ``Avg." denotes the mean AUC score.}
\label{TableR}%
\end{table}%

\subsection{Ablation Study}
In this subsection, we assess the contribution of different components in our method to the generalization performance. Experiments follow the cross-domain setting from \cite{chen2022self,liu2021spatial} using FF++ (HQ) as training and CDFv2 and DFDCP as the test sets. Ablation studies on the effectiveness of the hyperparameters in Eq. \ref{eq.21}, as well as the impact of $t$, $k$, $p_{\mathcal{A}}$, and the spectral analysis of the Fo-Mixup augmentation, are all provided in the \textbf{supplementary material}.

\vspace{1mm}
\noindent\textbf{Effectiveness of Fo-Mixup Augmentation.} To assess the impact of Fo-Mixup augmentation and confidence sampling, we deactivate them individually and together. In our experiments, the baseline model is the ResNet-34 network \cite{he2016deep} which is trained without the proposed CR. As reported in Table \ref{Table5}, Fo-Mixup augmentation significantly enhances the generalization performance compared to the baseline FreqDebias-0, both with (FreqDebias-3) and without (FreqDebias-1) the proposed loss functions. In FreqDebias-1, the model is trained with Fo-Mixup augmentation and standard CE loss function, highlighting the standalone effectiveness of Fo-Mixup augmentation. FreqDebias-3 demonstrates its impact when combined with the proposed loss functions. In FreqDebias-4, we evaluate confidence sampling by comparing it to FreqDebias-3, where all synthesized samples are used without sampling. The results demonstrate that confidence sampling enhances generalization by identifying the informative samples. 
\vspace{1mm}
\begin{table}
\center

\resizebox{0.46\textwidth}{!}{%
\begin{tabular}{l|ccccc|cc|cc}
\hline \hline \multirow{2}{*}{ Model } &\multicolumn{5}{c|}{ Components } &  \multicolumn{2}{c|}{CDFv2}&  \multicolumn{2}{c}{DFDCP}\\
\cline { 2 - 10} &DA & $L_{ {att }}$ & $L_{ {sphere }}$ & Fo-Mixup &   CS  &AUC &EER& AUC&EER \\
\hline 
FreqDebias-0 &$\checkmark$ & & & &   & 69.2 & 36.0& 61.3&40.7 \\
{FreqDebias-1} && &  &{$\checkmark$}  &  &   74.8 & 31.3& 68.1&36.0 \\

FreqDebias-2 &$\checkmark$& $\checkmark$ & $\checkmark$ &  &  &   77.3& 29.4& 73.6&32.2\\
FreqDebias-3 &$\checkmark$& $\checkmark$ & $\checkmark$ & $\checkmark$ &   & 81.7& 25.5 & 80.1&27.7 \\
FreqDebias-4 &$\checkmark$& $\checkmark$ & $\checkmark$ & $\checkmark$ &$\checkmark$   & 83.6& 23.9 & 82.4&26.1 \\

 \hline
      FreqDebias-5 & $\checkmark$& & $\checkmark$ & $\checkmark$ & $\checkmark$ & 79.1
 & 27.7 & 77.4&29.5 \\
      FreqDebias-6&$\checkmark$& $\checkmark$ & & $\checkmark$ &  $\checkmark$ & 80.4
 & 26.5& 78.2&28.9  \\

 \hline \multirow{2}{*}{  } &\multicolumn{7}{c|}{ Components } &  \multicolumn{2}{c}{CDFv2}\\
 \cline {2 - 10}&DA & $L_{ {att }}$ & $L_{ {sphere }}$ & Fo-Mixup &   CS  & AM \cite{xu2021fourier}&AS \cite{xu2021fourier}&AUC &EER \\
\hline

 FreqDebias-7 &$\checkmark$& $\checkmark$ & $\checkmark$ &  & $\checkmark$  & $\checkmark$ &   & 78.5&28.2 \\
 
 FreqDebias-8 &$\checkmark$& $\checkmark$ & $\checkmark$ & &  $\checkmark$ &  & $\checkmark$  & 76.0&30.3 \\

 \hline \multirow{2}{*}{  } &\multicolumn{7}{c|}{ Components } &  \multicolumn{2}{c}{CDFv2}\\
 \cline {2 - 10}&DA & $L_{ {att }}$ & $L_{ {sphere }}$ & Fo-Mixup &   CS  & \multicolumn{2}{c|}{ $\lambda$ }&AUC &EER \\
\hline

 FreqDebias-9 &$\checkmark$& $\checkmark$ & $\checkmark$ & $\checkmark$ & $\checkmark$  &  \multicolumn{2}{c|}{ 0.80 } & 82.1 & 25.2\\

 FreqDebias-10 &$\checkmark$& $\checkmark$ & $\checkmark$ & $\checkmark$ & $\checkmark$  &  \multicolumn{2}{c|}{ 0.65 } & 82.9 & 24.5\\

 FreqDebias-11 &$\checkmark$& $\checkmark$ & $\checkmark$ & $\checkmark$ & $\checkmark$  &  \multicolumn{2}{c|}{  0.50 } & 83.6 & 23.9\\

  FreqDebias-12 &$\checkmark$& $\checkmark$ & $\checkmark$ & $\checkmark$ & $\checkmark$  &  \multicolumn{2}{c|}{  0.35 } & 82.4 & 24.9\\
  


 
\hline
\hline
\end{tabular}}

\caption{Ablation study on the Fo-Mixup augmentation, confidence sampling (CS), and standard data augmentation (DA). All detectors are trained on the FF++ (HQ) dataset \cite{rossler2019faceforensics}, and evaluated on other datasets using the \textbf{frame-level} AUC metric.}
\label{Table5}%
\end{table}

\noindent\textbf{Fo-Mixup vs. Frequency Augmentations.} To demonstrate the superiority of Fo-Mixup, we replace it with traditional frequency-based augmentations such as amplitude swap (AS) and amplitude mix (AM) from \cite{xu2021fourier}, as shown in Table \ref{Table5}. Comparative evaluations in FreqDebias-7 and FreqDebias-8 verify Fo-Mixup as an effective augmentation for detecting face forgeries. AM indiscriminately modifies all frequency components, neglecting key frequencies, while AS aggressively swaps amplitude spectrums. In contrast, Fo-Mixup targets dominant frequency components, enhancing the detector's exposure to critical frequency variations.

\vspace{1mm}
\noindent\textbf{Effectiveness of Consistency Regularization.} In this experiment, we keep the Fo-Mixup component and explore the effect of the proposed attention and hyperspherical CRs. As shown in Table \ref{Table5}, our FreqDebias framework shows significant performance drops in AUC on the CDFv2 dataset, approximately $4.5\%$ and $3.2\%$, when the $L_{att}$ and $L_{sphere}$ regularizations are deactivated (in FreqDebias-5 and FreqDebias-6, respectively). We do not ablate the effect of ${L}_{cls\_sphere}$ since removing ${L}_{cls\_sphere}$ from the total loss function no longer optimizes the vMF classifier with learnable $\kappa$ and ${{\mu}}$, which are also used in ${L}_{sphere}$.\vspace{1mm}

\noindent\textbf{Choice of $\lambda$ value.} We investigate the impact of different $\lambda$ values, representing the percentage of selected samples, on generalization performance. We test $\lambda$ values of 0.35, 0.50, 0.65, and 0.80 across FreqDebias-12 to FreqDebias-9 experiments in Table \ref{Table5}. Our findings reveal that while a moderate lambda value of 0.50 is beneficial, extreme values (too high or too low) cause performance degradation. With $\lambda=0.80$ and $\lambda=0.35$, the model shows a significant drop in AUC, indicating overfitting or insufficient augmentation. Therefore, we use $\lambda=0.50$ in the remaining experiments to balance Fo-Mixup’s effectiveness without overfitting. \vspace{1mm}

\begin{table}
\center
\resizebox{0.32\textwidth}{!}{%
\begin{tabular}{l|cc|cc}
\hline \hline \multirow{2}{*}{ Model } &  \multicolumn{2}{c|}{CDFv2} &  \multicolumn{2}{c}{DFDCP}\\
\cline { 2 - 5} & AUC & EER & AUC & EER \\
\hline 
     
        FreqDebias-13 (ResNet-50) & ${83.9}$ & ${23.7}$ & ${82.9}$ & ${25.7}$  \\
        FreqDebias-14 (ConvNeXt) & ${85.1}$ & ${22.6}$ & ${83.4}$ & ${25.3}$  \\



\hline
\hline
\end{tabular}}
\caption{Ablation study on different backbones. All detectors are trained on the FF++ (HQ) dataset \cite{rossler2019faceforensics}, and evaluated on other datasets using the \textbf{frame-level} AUC metric.}
\label{Table_ab3}%
\end{table}

\noindent\textbf{Effectiveness of Backbone.} To evaluate the dependency of our performance on the ResNet-34 backbone, we substitute ResNet-34 with the ConvNeXt \cite{liu2022convnet} and ResNet-50 backbones. The performance of our FreqDebias framework employing the ResNet-50 and ConvNeXt backbones (FreqDebias-13 and FreqDebias-14 in Table \ref{Table_ab3}) surpasses the ResNet-34 backbone. This substantiates the adaptability of our framework for generalized forgery detection with different backbone choices.

\subsection{Visualization Analysis}
\begin{figure}[t] 
\centering
    \includegraphics[width=0.47\textwidth]{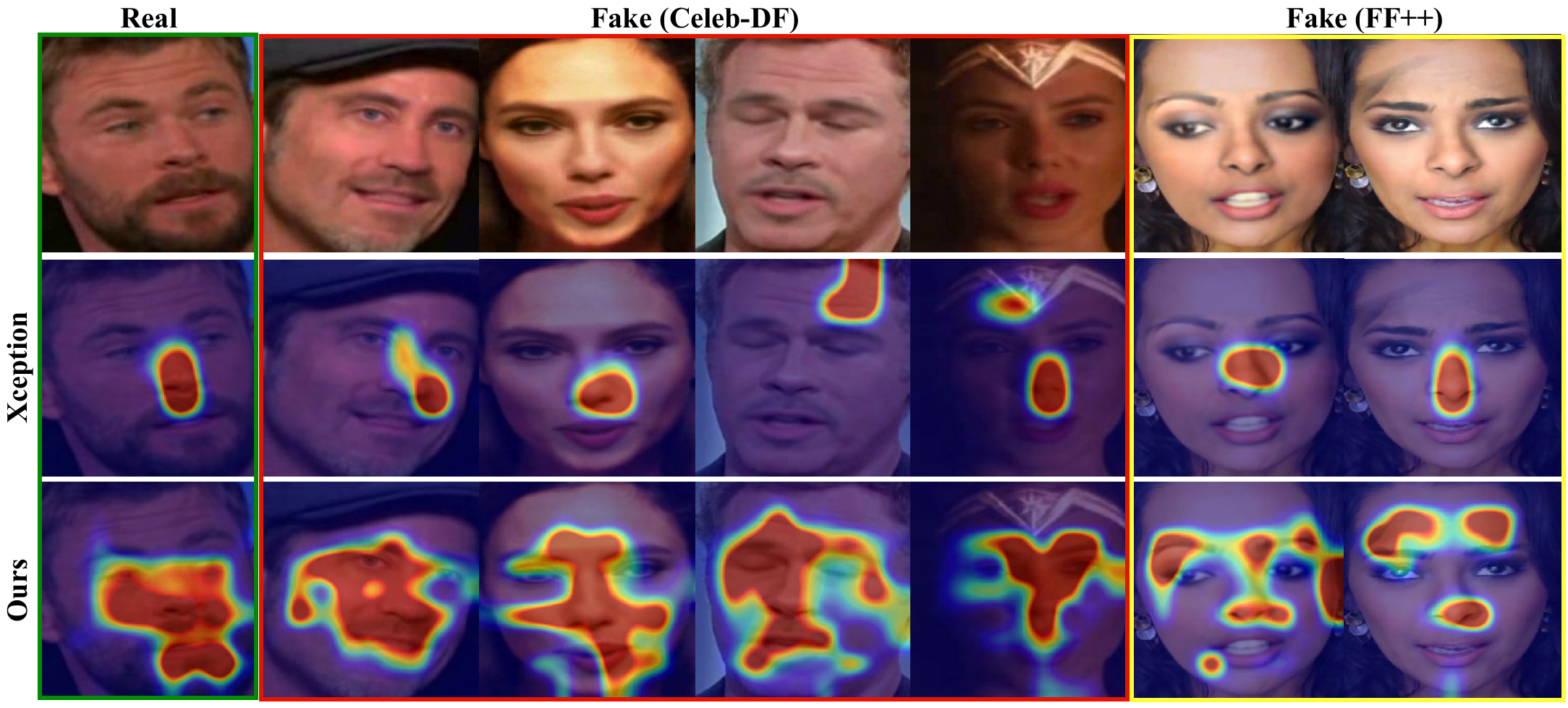}
    \caption{Visualization of the saliency maps for the Xception model \cite{rossler2019faceforensics} and the proposed FreqDebias framework.}
    \label{fig_cam}
\end{figure}

\noindent\textbf{Analysis of Saliency Map.} To localize and visualize critical facial regions in unseen forgeries, we apply CAM to our FreqDebias framework and the Xception model \cite{rossler2019faceforensics} trained on FF++. As seen in Fig. \ref{fig_cam}, Xception overfits to central facial regions, while FreqDebias effectively captures comprehensive regions, distinguishing between various forgery types and directing attention to respective artifacts, like eyebrow artifacts in FS forgeries. For instance, in the FS forgery (last column, Fig. \ref{fig_cam}), the FreqDebias framework focuses on artifacts in the eyebrow area.

\begin{figure}[t]
\centering
    \includegraphics[width=0.4\textwidth]{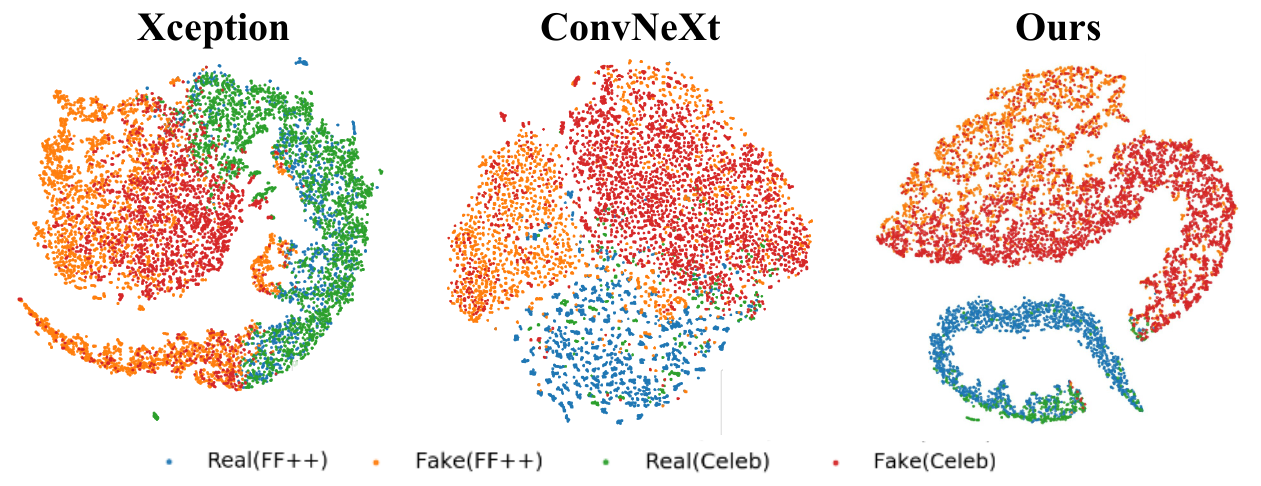}
    \caption{t-SNE illustration comparing the proposed FreqDebias framework with ConvNeXt \cite{liu2022convnet} and Xception \cite{rossler2019faceforensics} models, all trained on the FF++ dataset.}
    \label{fig_TSNE}
\end{figure}

\begin{figure}[t] 
\centering
   \includegraphics[width=0.47\textwidth]{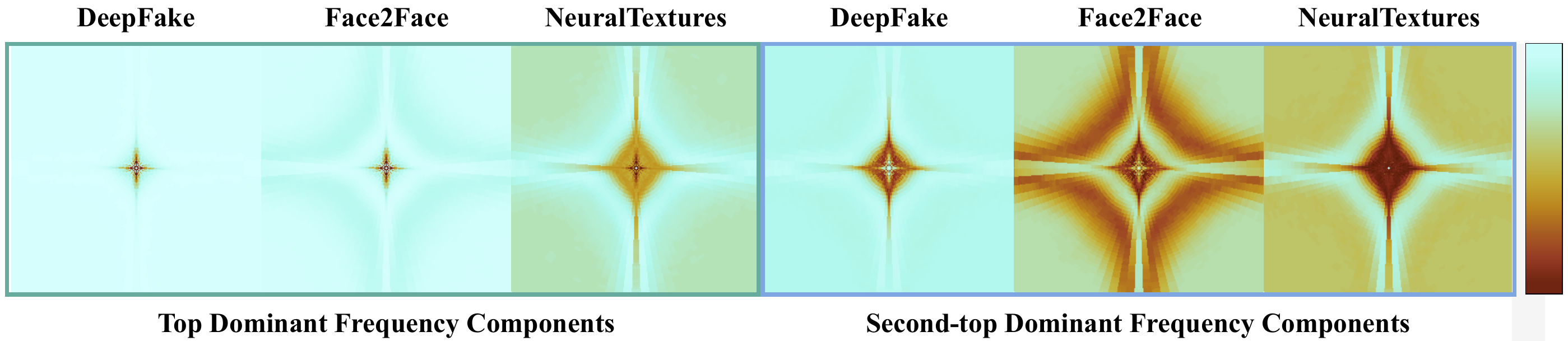}
    \caption{Heatmap of average dominant frequency components detected
by Fo-Mixup using a standard deepfake detector.}  
    \label{mask2}
\end{figure}

\vspace{1mm}
\noindent\textbf{Analysis of Feature Space.} We employ t-SNE \cite{van2008visualizing} to visualize the feature distribution learned by our FreqDebias framework. As shown in Fig. \ref{fig_TSNE}, compared to Xception \cite{rossler2019faceforensics} and ConvNeXt \cite{liu2022convnet}, FreqDebias better clusters real samples with class-wise separation and fewer outliers. While it is not explicitly trained to classify forgery types, the directional constraints of the vMF distribution enable FreqDebias to capture underlying structures, leading to partial separation despite some overlap at finer scales. Moreover, FreqDebias shows superior discrimination between real and forgery samples with enhanced generalization.\vspace{1mm}

\noindent\textbf{Illustration of the Dominant Frequency Components.} Fig. \ref{mask2} displays heatmaps of the average dominant frequency components heavily relied upon by a standard deepfake detector across 30,000 forgery images from FF++ \cite{rossler2019faceforensics}. For each forgery type, 10,000 images with similar identities are selected. In DeepFake \cite{Deepfakes_2019} forgeries, the detector focuses solely on very low-frequency bands, ignoring mid and high-frequency bands. Face2Face \cite{thies2016face2face} forgeries, however, also rely on mid-frequency bands. NeuralTextures \cite{thies2019deferred} forgeries differ further, as the dominant frequency components also span high-frequency bands.

\section{Conclusion}

This paper presents FreqDebias, a novel frequency debiasing framework that mitigates spectral bias in deepfake detection to enhance generalization ability. FreqDebias includes two key strategies: (1) Forgery Mixup (Fo-Mixup), an augmentation technique that broadens the detector’s exposure to a diversified frequency spectrum, and (2) dual consistency regularization (CR), which explicitly regulates learning across frequency components via local and global consistency. Extensive experiments show that addressing spectral bias with FreqDebias achieves state-of-the-art generalization in in-domain and cross-domain settings.
\section*{Acknowledgement}

This material is based upon work supported by the National Science Foundation under Grant Numbers CNS-2232048, and CNS-2204445. This research used in part resources on the Palmetto Cluster at Clemson University under National Science Foundation awards MRI 1228312, II NEW 1405767, MRI 1725573, and MRI 2018069. The views expressed in this article do not necessarily represent the views of NSF or the United States government.

{
    \small
    \bibliographystyle{ieeenat_fullname}
    \bibliography{main}
}


\end{document}